%% file: camera-ready.tex
\documentclass[letterpaper]{article} 
\usepackage{aaai24}  
\usepackage{times}  
\usepackage{helvet}  
\usepackage{courier}  
\usepackage[hyphens]{url}  
\usepackage{graphicx} 
\usepackage{natbib}  
\usepackage{caption} 
\usepackage{amsmath}
\usepackage{amsthm}
\usepackage{booktabs}
\usepackage{multirow}
\usepackage{diagbox}
\usepackage{microtype}
\usepackage{graphicx}
\usepackage{subfigure}
\usepackage[ruled,linesnumbered,vlined]{algorithm2e}
\usepackage{amsmath}
\usepackage{amssymb}
\usepackage{amsthm}
\usepackage{color}
\usepackage{diagbox}
\usepackage{multicol}
\usepackage{multirow}
\usepackage{threeparttable}
\input{math_commands.tex}

\title{Deep Active Learning with Noise Stability}
\author{
    Xingjian Li\textsuperscript{\rm 1}\equalcontrib,
    Pengkun Yang\textsuperscript{\rm 2}\equalcontrib,
    Yangcheng Gu\textsuperscript{\rm 3},
    Xueying Zhan\textsuperscript{\rm 1},
    Tianyang Wang\textsuperscript{\rm 4},\\
    Min Xu\textsuperscript{\rm 1}\thanks{Corresponding author},
    Chengzhong Xu\textsuperscript{\rm 5}
}
\affiliations{
    \textsuperscript{\rm 1}Computational Biology Department, Carnegie Mellon University \\
    \textsuperscript{\rm 2}Center for Statistical Science, Tsinghua University \\
    \textsuperscript{\rm 3}School of Software, Tsinghua University \\
    \textsuperscript{\rm 4}Department of Computer Science, University of Alabama at Birmingham\\
    \textsuperscript{\rm 5}State Key Lab of IOTSC, University of Macau\\
    lixj04@gmail.com,
    yangpengkun@tsinghua.edu.cn, 
    guyc19@mails.tsinghua.edu.cn\\
    sinezhan17@gmail.com,
    tw2@uab.edu,
    mxu1@cs.cmu.edu,
    czxu@um.edu.mo

}

\begin{document}

\maketitle

\begin{abstract}
Uncertainty estimation for unlabeled data is crucial to active learning. With a deep neural network employed as the backbone model, the data selection process is highly challenging due to the potential over-confidence of the model inference. Existing methods resort to special learning fashions (e.g. adversarial) or auxiliary models to address this challenge. This tends to result in complex and inefficient pipelines, which would render the methods impractical. In this work, we propose a novel algorithm that leverages noise stability to estimate data uncertainty. The key idea is to measure the output derivation from the original observation when the model parameters are randomly perturbed by noise. We provide theoretical analyses by leveraging the small Gaussian noise theory and demonstrate that our method favors a subset with large and diverse gradients. Our method is generally applicable in various tasks, including computer vision, natural language processing, and structural data analysis. It achieves competitive performance compared against state-of-the-art active learning baselines.
\end{abstract}

\input{sections/1_introduction}
\input{sections/5_relatedwork}
\input{sections/2_algorithm}
\input{sections/3_experiment}

\input{sections/4_discussion}
\input{sections/6_conclusion}

\bibliography{aaai24}

\newpage
\input{sections/9_appendix}

\end{document}

%% file: math_commands.tex
\usepackage{amsmath,amsfonts,bm}

\def\eqref#1{equation~\ref{#1}}

\def\1{\bm{1}}

\DeclareMathAlphabet{\mathsfit}{\encodingdefault}{\sfdefault}{m}{sl}
\SetMathAlphabet{\mathsfit}{bold}{\encodingdefault}{\sfdefault}{bx}{n}

\DeclareMathOperator*{\argmax}{arg\,max}

%% file: sections/1_introduction.tex
\section{Introduction}
The success of training a deep neural network highly depends on a huge amount of labeled data. 
However, much data is unlabeled in the era of big data 
, and data annotation cost could be extremely high, such as in medical area \cite{gorriz2017cost, konyushkova2017learning}. Due to limited annotation budget, it might be more feasible to annotate a portion of the given data. 
Then a question can be naturally raised: which data is deserved for being annotated? 

Active learning was proposed to solve this question. 
One of the main ideas 
is to select the most uncertain/informative data from an unlabeled pool for annotation. 
The newly annotated data is then used to train a task model in supervised fashion. Nevertheless, uncertainty estimation with deep neural networks (DNNs) remains challenging,
due to the potential over-confidence of deep models~\cite{ren2021survey}. That is, considering classification problems, the primitive softmax output (as a score) does not necessarily reflect a reliable uncertainty or confidence of the prediction. Several methods have been proposed to address this challenge. For example, deep Bayesian methods~\cite{gal2016dropout} provide a more accurate estimate of uncertainty (e.g. in terms of entropy or mutual information) with the posterior predictive probability over the model parameters. 
Many other methods propose to employ auxiliary models for a better estimate, such as query-by-committee (QBC)~\cite{1997Selective,gorriz2017cost},  Variational Auto-Encoder~\cite{sinha2019variational}, adversarial learning~\cite{ducoffe2018adversarial,mayer2020adversarial} and Graph Convolutional Network~\cite{Caramalau_2021_CVPR}.

Despite the advances in uncertainty estimation, 
they rely on additional models or ad-hoc learning fashions, highly limiting their applications. Furthermore, many of them fail in considering the diversity of the selected data, throwing a risk of selecting redundant or similar examples in practical scenarios of batch acquisition. Motivated to address these issues, our work aims to design a more general and effective active learning algorithm. 

In this work, we study the problem of deep active learning from a different perspective named \emph{noise stability}. 
Given a well-trained model and a data example, we are interested in how the model behavior changes when a slight perturbation is imposed on the model parameters. Specifically, we observe the deviation on the deep features of DNNs. A data example which is sensitive to such perturbations, is regarded as uncertain under the current model. The sensitivity can be measured by the magnitude of the feature deviation. Furthermore, the deviation direction implies common and different patterns among examples, which provides meaningful information to encourage diversity. Imagine the ideal case that, when the network perturbation causes the ablation of recognizing a specific pattern, examples that highly depend on this pattern will have more changes than others on corresponding elements of the feature vector\footnote{In real cases, a random perturbation leads to a complex change in model parameters, e.g. some extracted patterns are affected more than others, which consequently results in much denser deviation directions among examples. }. 

In light of the aforementioned intuitions, we design a novel deep active learning algorithm by 
measuring the feature deviation from the original observation, 
when imposing a small noise on the model parameters. The magnitude of the deviation acts as a metric of uncertainty, and the direction is utilized to constitute a diverse subset of selected unlabeled examples. 
We provide a theoretical analysis under the condition that the noise magnitude is small. By imposing standard multivariate Gaussian noise to the parameters, we prove that the induced deviation yields a sort of \emph{projected gradient}. The expectation of the magnitude of this gradient is equivalent to the gradient norm w.r.t the model parameters, whereas the direction of this gradient serves as a proxy of the direction of the original gradient.

Our method is easy to implement and free of data-dependent designs or  auxiliary models. Therefore, it can be exploited in classification, regression and more complex tasks such as semantic segmentation. It is also applicable to various data domains such as image, natural language and structural data. We 
evaluate our method on multiple tasks, including (1) image classification on MNIST~\cite{lecun1998gradient}, Cifar10 \cite{krizhevsky2009learning}, SVHN \cite{svhn}, Cifar100 \cite{krizhevsky2009learning}, and Caltech101 \cite{caltech}; (2) regression on the Ames Housing dataset~\cite{de2011ames}; (3) semantic segmentation on Cityscapes \cite{cityscapes};  and (4) natural language processing on MRPC~\cite{dolan2005automatically}. Our method is shown to exceed or be comparable to the state-of-the-art baselines on all these tasks.

The contributions of this work are summarized as follows. 
\begin{enumerate}
    \item We propose a novel deep active learning method, namely \emph{NoiseStability}, to select unlabeled data for annotation. It is free of any auxiliary models or customized training fashions. We conduct extensive experiments on diverse tasks and data types to demonstrate the general applicability and superior performance of our method. 
    \item We prove that selecting unlabeled data w.r.t. noise stability is theoretically equivalent to selecting data according to randomly projected gradients. Conditioned on standard Gaussian noise, the induced feature deviation exhibits decent properties in both estimating uncertainty and ensuring diversity. 
    \item Connections of our method with predictive variance reduction and gradient-based active learning methods are discussed in detail. 

\end{enumerate}

%% file: sections/5_relatedwork.tex
\section{Related Work}

\subsection{Active Learning}
Active learning, aiming at determining which data to be annotated, has been a long-term open research topic in the community. Among existing literature, two most important criterion for data selection are uncertainty and diversity. 

\noindent \textbf{Uncertainty Estimation.} Traditional active learning methods take the data, which leads to a \emph{low predictive confidence}, as most informative for further learning. Typical strategies select data of which predictions are close to the decision boundary~\cite{tong2001support,dasgupta2008hierarchical,vzliobaite2013active} or inconsistent among different hypotheses (known as version-space based approach) ~\cite{cohn1994improving,freund1997selective}. On over-confident DNNs, to tackle the challenge of incredible probabilistic predictions, these ideas are implemented through deep Bayesian approximation with MC-Dropout~\cite{gal2016dropout}, query-by-committee (QBC) with multiple-model training~\cite{gorriz2017cost} or adversarial training~\cite{ducoffe2018adversarial,mayer2020adversarial}. 

Some other approaches aim to choose spaces which have a \emph{low data density}. For example, \cite{sinha2019variational,zhang2020state} adopt Variational Autoencoder (VAE) to model data distributions, and then select the data that less likely lies in the labeled set in an adversarial fashion. ~\cite{Caramalau_2021_CVPR} uses Graph Convolutional Networks (GCN) to characterize the relationship among examples and distinguish those unlabeled examples which are sufficiently different from labelled ones. 

Another typical idea focuses on possible model changes caused by adding candidate examples, and favors those \emph{highly influential data}. For example, \cite{CohnD.A1996ALwS} introduces the goal of selecting a new example that is expected to minimize the predictive variance on existing labeled examples. \cite{yoo2019learning} incorporates a light-weight module to learn the predictive error of unlabeled examples, namely Learning Loss for Active Learning (LL4AL).  \cite{liu2021influence,wang2022boosting} instead suggest leveraging the influence function~\cite{koh2017understanding} to estimate the potential model change. 

For differentiable functions such as DNNs, the Expectation Gradient Length (EGL) has been shown  an effective criteria to measure the influence (to the model) of a given example~\cite{settles2008analysis}, which motivates active learning applications in specific domains, such as speech recognition~\cite{huang2016active} and text classification~\cite{zhang2017active}. However, since labels are not available in active learning, these methods have to resort to psuedo labels or a marginalization over the label space. 

\noindent \textbf{Diversification.} 
In a more realistic setting where a batch of examples are selected at one time, active learning algorithms should avoid redundancy among selected examples. 
To tackle this problem, distribution-based methods~\cite{nguyen2004active,yang2015multi,elhamifar2013convex, hasan2015context} have been designed. Inspired by the theory of set cover, a recent state-of-the-art method Coreset~\cite{sener2018active} proposes to select examples with diverse representations. \cite{loquercio2020general} further investigates the relevance between data diversity and  uncertainty. 

Moreover, several specific uncertainty-based approaches successfully take diversity into consideration. VAAL \cite{sinha2019variational} selects diverse examples by choosing data from different regions of VAE's latent space. CoreGCN~\cite{Caramalau_2021_CVPR} proposes to select diverse data through graph embeddings. Cluster-Margin~\cite{citovsky2021batch} leverages Hierarchical Agglomerative Clustering (HAC) to diversify a batch of most uncertain examples.

For gradient-based methods, enhancing diversity is a  non-trivial objective. This is because when comparing gradient directions, one has to deal with the extremely huge dimension of the gradients (w.r.t. model parameters).
A recent state-of-the-art method BADGE~\cite{ash2019deep}  proposes a practical solution by using the parameters of the last neural layer.

\subsection{Posterior Approximation for Bayesian Neural Network}
The Bayesian Neural Network (BNN) aims to quantify the uncertainty by considering the probabilistic distribution of parameters, rather than  deterministic ones.  In terms of predictive variance reduction, our work is  relevant to the linearised Laplace approximation of the BNN posterior~\citep{immer2021scalable,daxberger2021bayesian} and BAIT~\citep{ash2021gone} which involves the
negative Hessian matrix.



Despite their theoretical relevance, the resulting algorithms are different: \cite{immer2021scalable} is based on the determinant of the Hessian matrix; the method in \cite{daxberger2021bayesian} involves the diagonal of inverse Hessian. Our method is more efficient and scalable, where choosing examples with larger  and diverse projected gradients serves as the cornerstone of our approach.

\subsection{Noise Stability for Supervised Learning}
Noise stability, particularly in terms of the output stability w.r.t. the input noise, has been studied in standard supervised learning. For example, \cite{bishop1995training} proposes to train a model with perturbed input examples as a regularizer to improve the model performance. \cite{Arora2018StrongerGB} shows that, the stability of each layer’s computation towards input noise, injected at lower layers, acts as a good indicator of the generalization bounds for DNNs. An intuition behind these achievements is that, a model (more exactly, its parameters) robust to perturbed input examples should perform better in recognizing unseen examples (e.g. test data). 

Inspired by the aforementioned intuition, in this work we explore how the robustness of model parameters impact data selection in active learning. In fact, this problem can be regarded as an inverse problem of the classical noise stability. In terms of the objective, our method pursues informative data which eventually leads to an enhanced model generalization.


%% file: sections/2_algorithm.tex
\section{Deep Active Learning with Noise Stability}
\subsection{Problem Definition}
Here we formulate the deep active learning problem. Given a pool of unlabeled data $\mathcal{U}$, and a labeled pool $\mathcal{L}$ which is initially empty. Active learning aims to select a portion of data from $\mathcal{U}$ depending on an annotation budget (e.g. select 2500 samples out of 50000 at a time). The selected data \{$X_N$\} is then annotated by human oracles (or equivalent), and added to the labeled pool. That is, $\mathcal{L} \leftarrow$ $\mathcal{L}$ $+$ \{$X_N$, $Y_N$\}, where \{$Y_N$\} is the new annotation for \{$X_N$\}. The unlabeled pool is then updated by removing the selected data: $\mathcal{U}$ $\leftarrow$ $\mathcal{U}$ $-$ \{$X_N$\}. The updated $\mathcal{L}$ is then used to train a task model 
composed of a feature extractor $f$
followed by a task head $g$.
$f$ and $g$ are parameterized by $\bm{\theta}$ and $\bm{\phi}$, respectively. We use $f(.;\bm{\theta})$ and $g(.;\bm{\phi})$ to denote the corresponding feed-forward operations of $f$ and $g$. 
Note that the feature extractor $f$ is normally a deep neural network 
of which parameters $\bm{\theta} \in \mathcal{R}^n$ are in high dimension. In contrast, the architecture of $g$ is 
arguably shallow,
e.g. a single fully connected layer for classification or regression problems.

\begin{algorithm}[t]
\small
\SetKwData{P_u}{P_u}\SetKwData{P_v}{P_v}\SetKwData{P_i}{P_i}\SetKwData{P_j}{P_j}
\SetKwData{T1}{t1}\SetKwData{T2}{t2} \SetKwData{E}{E}
\SetKwData{Left}{left}\SetKwData{This}{this}\SetKwData{Up}{up}
\SetKwInOut{Input}{Input}\SetKwInOut{Output}{Output}
\Input{
$\mathcal{T}$: random initialized neural network, $\mathcal{U}$: unlabeled pool of training data, $\mathcal{L}$: labeled pool of training data;
}
\Output{
$\mathcal{L}$: updated labeled pool;
}

\Begin{
train $\mathcal{T} = g \circ f$ with $\mathcal{L}$, obtaining the current parameters $\bm{\theta},\bm{\phi}$; \\

\For{$k \leftarrow 1$ \KwTo $K$}{
    sample a random direction $\bm{u}^{(k)}$ ;\\
    get $\Delta\bm{\theta}^{(k)} = \zeta \|\bm{\theta}\|_2 \bm{u}^{(k)}$; \\
} 
\For{every $\bm{x}$ in $\mathcal{U}$}{ 
    \For{$k \leftarrow 1$ \KwTo $K$}{
         $\Delta z(\bm{x})^{(k)} = f(\bm{x};\bm{\theta} +  \Delta \bm{\theta}^{(k)}) - f(\bm{x};\bm{\theta})$; \\
    } 
    get $\widehat{\Delta z}(\bm{x})$ by concatenating all $\Delta z(\bm{x})^{(k)}$ for $k=1,2,...,K$; \\
} 
compute $N$ centers $\{c_j\}_{j=1}^{N}$ w.r.t. $\{\widehat{\Delta z}(\bm{x})|\bm{x} \in \mathcal{U}\}$ by greedy k-center algorithm; \\
select samples whose indexes are $\{c_j\}_{j=1}^{N}$ in $\mathcal{U}$ as $\{X_N\}$, and obtain their labels $\{Y_N\}$; \\
update $\mathcal{L}$ with $\mathcal{L}=\mathcal{L}\bigcup \{X_N,Y_N\}$; \\
return $\mathcal{L}$ ;
}
\caption{Active learning with Noise Stability.}
\label{algo}
\end{algorithm}

\subsection{Uncertainty Estimation with Noise Stability}
We aim at investigating how intermediate representations change when a small perturbation is imposed on model parameters.
Specifically, given $x$ as an input, we consider the output of the feature extractor, i.e. $z = f(\bm{x};\bm{\theta}) \in \mathcal{R}^d$.  Let $\Delta \bm{\theta} \in \mathcal{R}^n$ denote the added noise (perturbation) and $\Delta \bm{\theta} = \zeta \|\bm{\theta}\|_2 \bm{u}$,
where $\zeta$ controls the relative magnitude of the noise and $\bm{u}$ is a random direction with the unit length.\footnote{In practice, $\bm{u}$ can be sampled from a standard multivariate Gaussian distribution and then normalized by its own $L^2$-norm.}  $\zeta$ should be relatively small (e.g. $10^{-3}$) in order to avoid a catastrophic perturbation to the clean model.
Then, for an unlabeled sample $\bm{x} \in X_U$, its feature deviation is
\begin{equation}
\Delta z(\bm{x}) = \frac{1}{\zeta \|\bm{\theta}\|_2} (f(\bm{x};\bm{\theta} + \Delta \bm{\theta}) - f(\bm{x};\bm{\theta})).
\end{equation}
For brevity, 
we assume $z$ and $\Delta z$ are both row vectors, 
and use $\Delta \bm{\theta}^{(k)}$ to denote the $k$-th sampled parameter noise. 
We perform $K$ samplings of $\Delta \bm{\theta}$ and concatenate all the resulting feature deviations as
\begin{equation}
\label{eq:main}
    \widehat{\Delta z}(\bm{x}) = \sqrt{\frac{n}{K}} (\Delta z(\bm{x})^{(1)}, \Delta z(\bm{x})^{(2)}, ..., \Delta z(\bm{x})^{(K)}), 
\end{equation}
where $\widehat{\Delta z}(\bm{x}) \in \mathcal{R}^{Kd}$ is the empirical result and each $\Delta z(\bm{x})^{(k)}$ corresponds to the feature deviation produced by the parameter noise $\Delta \bm{\theta}^{(k)}$. 

Here we adopt the empirical feature deviation $\widehat{\Delta z}(\bm{x})$ as the representation of each unlabeled example $\bm{x}$. 
The rationale lies in two aspects.
Firstly, in active learning, we favor data with higher uncertainty 
w.r.t. the current learned model. Such data are usually hard examples that tend to be sensitive to small parameter perturbations. Therefore, the 
length of $\widehat{\Delta z}(\bm{x})$ can be used as a criterion for uncertainty estimation. Secondly, the direction of $\widehat{\Delta z}(\bm{x})$ implies potential characteristics of an example $\bm{x}$, that is, 
which learned feature components are sensitive (or not robust) to a parameter perturbation. 
Diverse currently non-robust features should be focused on, in order to learn more comprehensive knowledge. 
Due to its relevance to data diversity, the direction of $\widehat{\Delta z}(\bm{x})$ can be utilized to select a batch of samples, where the diversity of selected examples is also essential to a good performance~\cite{sener2018active,kirsch2019batchbald}. 

Considering both  magnitude and direction, we use Gonzalez's greedy implementation~\cite{gonzalez1985clustering} of the k-center algorithm for diverse subset selection. As pointed out in BADGE~\cite{ash2019deep}, a diverse subset will naturally prefer examples with larger magnitudes as they tend to be far away from each other if measured in a Euclidean space. The pipeline of our method is summarized in Algorithm~\ref{algo}.


\section{Theoretical Understandings}
In this section, we present theoretical understandings of our method for deep active learning. Our main conclusions are summarized as follows. 

\begin{itemize}
    \item When the noise magnitude $\zeta\|\bm{\theta}\|$ is sufficiently small, for a given example $\bm{x}$, the expectation of $\|\widehat{\Delta z}(\bm{x})\|_2$ is the Frobenius norm of the Jacobian of $f$ w.r.t. its parameters $\bm{\theta}$, when proper prerequisites are satisfied for $\bm{u}$.
    \item Considering the 1-d case $f: X \to \mathcal{R}$, the suggested $\widehat{\Delta z}(\bm{x})$ can be regarded as a projected gradient, which has a much lower dimension compared to the original gradient $\nabla_{\bm{\theta}}f$. 
    With the projected gradient being employed, the spatial relationships (e.g. Euclidean distance and inner product) among examples remain the same as if the original gradient is being used, 
    indicating $\widehat{\Delta z}(\bm{x})$ a reasonable proxy of the original gradient, when considering diversity.
    
\end{itemize}

\subsection{Noise Stability as Jacobian Norm}
Let $f(\bm{x};\bm{\theta})$ be differentiable (w.r.t. $\bm{\theta}$) at $\bm{\theta} \in \mathcal{R}^n$ given $\bm{x}$ as an input. When the imposed noise $\Delta \bm{\theta} = \zeta \|\bm{\theta}\|_2 \bm{u}$ has a sufficiently small magnitude, i.e. in the limit as $\zeta \to 0$, we use the first-order Taylor expansion to represent the perturbed output $f(\bm{x};\bm{\theta}+ \Delta \bm{\theta})$ as
\begin{equation}
    f(\bm{x};\bm{\theta}+\Delta \bm{\theta}) = f(\bm{x};\bm{\theta})+ \mathrm{J}_{\bm{\theta}}(\bm{x};\bm{\theta}) \Delta \bm{\theta} + o(\zeta \|\bm{\theta}\|_2),
\label{eq:taylor}
\end{equation}
where $\mathrm{J}_{\bm{\theta}}(\bm{x};\bm{\theta})$ denotes the Jacobian matrix of $f$ w.r.t the parameters $\bm{\theta}$ as $\mathrm{J}_{\bm{\theta}}(\bm{x};\bm{\theta})_{(i,j)} = \partial f(\bm{x};\bm{\theta})_{(i)} / \partial \bm{\theta}_{(j)}$. By applying the Taylor approximation, each block $\Delta z(\bm{x})^{(k)}$ can be formulated as
\begin{equation}
    \Delta z(\bm{x})^{(k)} =  \mathrm{J}_{\bm{\theta}}(\bm{x};\bm{\theta}) \bm{u}^{(k)} +  o(1).
\label{eq:zk}
\end{equation}
By omitting the higher order term $o(1)$, it results in the output deviation magnitude to be

\begin{equation}
    \|\widehat{\Delta z}(\bm{x})\|_2^2 = \frac{n}{K} \sum_{k=1}^{K} \|\mathrm{J}_{\bm{\theta}}(\bm{x};\bm{\theta}) \bm{u}^{(k)}\|_2^2.
\label{eq:taylor2}
\end{equation}
Conditioned on the elements of $\bm{u}^{(k)}$ are zero-mean and independent of each other, it can be derived that the expectation of the data selection metric (i.e. $\|\widehat{\Delta z}(\bm{x})\|_2$) is equivalent to the Frobenius norm of Jacobian as

\begin{equation}
    \underset{\bm{u}}{\mathbb{E}}
    \|\widehat{\Delta z}(\bm{x})\|_2^2 =  \|\mathrm{J}_{\bm{\theta}}(\bm{x};\bm{\theta})\|_F^2.
\label{eq:jacobian}
\end{equation}
We put a detailed proof of Eq~(\ref{eq:jacobian}) in Appendix A.1  using the Standard multivariate Gaussian distribution as a demonstration. In fact, Eq~(\ref{eq:jacobian}) 
is satisfied for any appropriate noise distributions with the aforementioned prerequisite of independence and zero mean. We also provide a quantified concentration inequality, showing that $\|\widehat{\Delta z}(\bm{x})\|_2^2$ can be close to the Jacobian norm with high probability in Appendix A.1.  

\subsection{Diversified Projected Gradients}
Since $\widehat{\Delta z}(\bm{x})$ indicates which directions (i.e. elements of the final feature) of an example is more sensitive to a specific parameter perturbation, it can be employed to select diverse examples to constitute a subset.
By applying the first order Taylor expansion, each $\widehat{\Delta z}(\bm{x})$ can be regarded as a \emph{projected gradient} when the output of $f$ is a scalar (here we ignore the Jacobian matrix for simplicity). That is, the original gradient $\nabla_{\bm{\theta}}f$ is transformed by random directions $\{\bm{u}^{(k)}\}_{k=1}^{K}$ to its projected format as shown in Eq (\ref{eq:zk}).

The benefit of diverse gradients has been discussed in existing literature. For example, a recent work~\cite{zhang2017determinantal} theoretically shows that, performing SGD with diversified gradients within a mini-batch helps lower the variance of stochastic gradients. BADGE~\cite{ash2019deep} selects diverse examples according to their last layer's gradients rather than considering that of the entire model. Here we show that, leveraging the expectation of the projected gradients $\{\widehat{\Delta z}(\bm{x}) | \bm{x} \in \mathcal{U} \}$ 
is equivalent to using $\{\nabla_{\bm{\theta}}f(\bm{x};\bm{\theta}) | \bm{x} \in \mathcal{U} \}$. Specifically, given two examples $\bm{x}_i$ and $\bm{x}_j$, the following two properties hold.

\noindent \textbf{Property 1.} \emph{Equivalence of Euclidean distance.} 
\begin{equation}
\label{eq:property1}
\underset{ \bm{u} }{\mathbb{E}}\|
\widehat{\Delta z}(\bm{x}_i) - \widehat{\Delta z}(\bm{x}_j)\|_2^2 =  \| \nabla_{\bm{\theta}}f(\bm{x}_i;\bm{\theta}) - \nabla_{\bm{\theta}}f(\bm{x}_j;\bm{\theta}) \|_2^2. 
\end{equation}

\noindent \textbf{Property 2.} \emph{Equivalence of inner product.} 
\begin{equation}
\label{eq:property2}
\underset{ \bm{u}}{\mathbb{E}}
\langle \widehat{\Delta z}(\bm{x}_i), \widehat{\Delta z}(\bm{x}_j) \rangle = \langle \nabla_{\bm{\theta}}f(\bm{x}_i;\bm{\theta}),  \nabla_{\bm{\theta}}f(\bm{x}_j;\bm{\theta}) \rangle. 
\end{equation}
We put a detailed proof of these properties in Appendix A.2. Similarly, we will also show that our estimated Euclidean distance and inner product are close to their respective expectations with high probability. 

The above analyses lead to the following conclusion: 
subset selection with $\{\widehat{\Delta z}(\bm{x}) | \bm{x} \in \mathcal{U} \}$ is approximately equivalent to that with  $\{\nabla_{\bm{\theta}}f(\bm{x};\bm{\theta}) | \bm{x} \in \mathcal{U} \}$, by using either distance-based (e.g. greedy-k-center) or similarity-based algorithms (e.g. k-DPP). 

\subsection{Connections to Variance Reduction}
\label{sec:variance}
A widely accepted criterion in active learning is to select examples which minimize the predictive variance on existing training data~\cite{CohnD.A1996ALwS}. This idea is established on the  theory of variance reduction~\citep{MacKayDavidJ.C1992IOFf,settles2009active} . For convenience,  
we reuse previous notations but assume that $\hat{y} = f(.;\bm{\theta})$ is a real-valued target function parameterized by $\bm{\theta}$ that results in $\hat{y} : \mathbb{R}^d \to \mathbb{R}$, and  
this assumption can naturally generalize to vector-valued functions. Given a labeled training set $\mathcal{L} = \{\bm{x}_i, y_i\}_{i=1}^{m}$, the learning objective is to minimize the Mean Squared Error (S) as $S^2 = \frac{1}{m} \sum_{i=1}^{m}(f(\bm{x}_i; \bm{\theta})-y_i)^2$.
According to the theory of variance estimation~\citep{alma991005548549706306}, we derive that, by adding a new sample $\tilde{\bm{x}}$, the reduction in the predictive variance of a training example $\bm{x}$  is:
\begin{equation}
\small
    \underset{\tilde{\bm{x}}}{\Delta} \sigma^2_{\hat{y}}(\bm{x}) =  \frac{\|\mathrm{J}_{\bm{\theta}}(\bm{x};\bm{\theta})^T \mathbf{A}_{S^2}^{-1}  \bm{u}(\tilde{\bm{x}})\|^2 \cdot \|\mathrm{J}_{\bm{\theta}}(\tilde{\bm{x}};\bm{\theta})\|^2}{ S^2+\| \bm{u}(\tilde{\bm{x}})^T \mathbf{A}_{S^2}^{-1}  \bm{u}(\tilde{\bm{x}}) \| \cdot \|\mathrm{J}_{\bm{\theta}}(\tilde{\bm{x}};\bm{\theta})\|^2},
\label{eq:var_change}
\end{equation}
where $\mathbf{A}_{S^2}$ refers to  the scaled Fisher Information Matrix: $\mathbf{A}_{S^2} = \frac{1}{S^2} \frac{\partial^2 S^2}{\partial \bm{\theta}^2}$, and $\bm{u}(\tilde{\bm{x}})$ of the unit length  denotes the direction of $\mathbf{J}_{\hat{y}}(\tilde{\bm{x}})$ as  $\bm{u}(\tilde{\bm{x}}) = \frac{\mathrm{J}_{\bm{\theta}}(\tilde{\bm{x}};\bm{\theta})}{\|\mathrm{J}_{\bm{\theta}}(\tilde{\bm{x}};\bm{\theta})\|^2}$.

Eq~(\ref{eq:var_change}) implies that, for any direction $\bm{u}(\tilde{\bm{x}})$, the optimal choice for maximizing $\underset{\tilde{\bm{x}}}{\Delta} \sigma^2_{\hat{y}}(\bm{x})$ is the ones with the largest $\|\mathrm{J}_{\bm{\theta}}(\tilde{\bm{x}};\bm{\theta})\|^2$. On the other hand, diverse directions $\bm{u}(\tilde{\bm{x}})$ tend to benefit different training examples more uniformly, which contributes to the overall learning objective. Details for the derivation of Eq~(\ref{eq:var_change}) is presented in Appendix A.3.

\subsection{Comparison with to BADGE}
Now we discuss more details comparing with BADGE~\cite{ash2019deep}, which is a state-of-the-art gradient-based active learning algorithm. In a multiclass classification problem with softmax activations, given a loss function $L$, BADGE yields the $i$-th block of the gradient of an example $\bm{x}$ as
\begin{equation}
    \nabla_{\bm{\phi}_i}L = (p_i-I(\hat{y}=i))f(\bm{x};\bm{\theta}),
\label{eq:badge}
\end{equation}
where $\bm{\phi}_i$ is the $i$-th row parameters of $\bm{\phi}$, and $p_i$ is the probability for label $i$, thus  $\hat{y}=\underset{i}{\argmax} \ p_i$. 
Though our method (based on noise stability) is highly relevant to gradients as shown in Eq~(\ref{eq:zk}), there are two essential differences with BADGE (as well as other related works).
\begin{itemize}
    \item Noise stability avoids explicitly approximating gradients, 
    and hence does not require the extensive computation of assuming and marginalizing pseudo labels.
    This issue can be fatal in complex scenarios. For example, in semantic segmentation (or speech recognition), the number of label assumptions increases exponentially with increase in the number of pixels (or timesteps). Therefore, it is infeasible to gather the gradients w.r.t. all possible labels. 
    \item In terms of final representations, noise stability 
    resorts to \emph{projected gradients} w.r.t. feature, while BADGE actually relies on \emph{feature itself} and probabilistic predictions. In terms of uncertainty estimation, the gradient magnitude $\|\nabla_{\bm{\theta}} f(\bm{x};\bm{\theta})\|_2$ is a more meaningful indicator than the magnitude of $f(\bm{x};\bm{\theta})$. Besides, in a regression problem, Eq~(\ref{eq:badge}) is meaningless due to the infeasibility of assuming a ground truth label and has to resort to the original feature only. In contrast, noise stability is still applicable, taking into account both uncertainty and diversity. 
    
\end{itemize}

\subsection{Implementation Tips}
In practice, for a deep neural network we can compute the perturbation $\widehat{\Delta z}(\bm{x})$ w.r.t. the final deep feature. However, for a  classification network, it is also feasible to employ the final predictive output to measure noise stability, since such an output can be regarded as a high-level semantic feature. Moreover, the predictive output usually compresses the primitive deep feature, thus resulting in a vector with a lower dimension. In applications with a large pool size and a high annotation budget, adopting the predictive output will naturally lead to an efficient implementation.
Therefore, we suggest to compute the deviation on the predictive output for classification tasks, and on the final deep feature for other tasks such as regression and semantic segmentation. 

%% file: sections/3_experiment.tex
\begin{figure*}[t]

    \centering
    \includegraphics[width=0.31\linewidth]{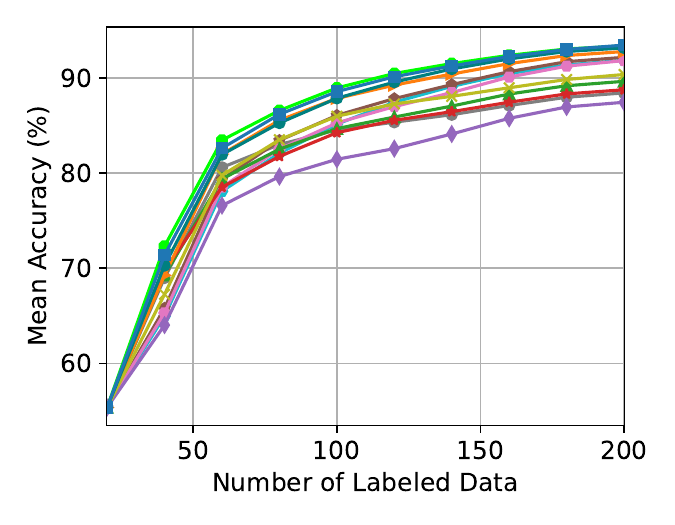}
    \includegraphics[width=0.31\linewidth]{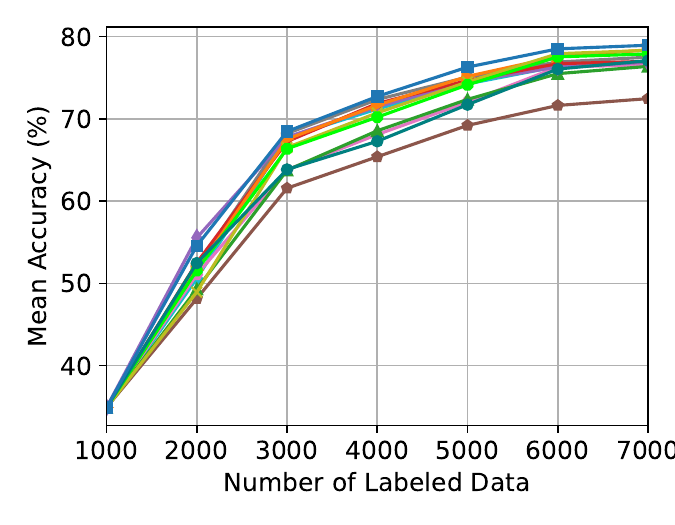}
    \includegraphics[width=0.31\linewidth]{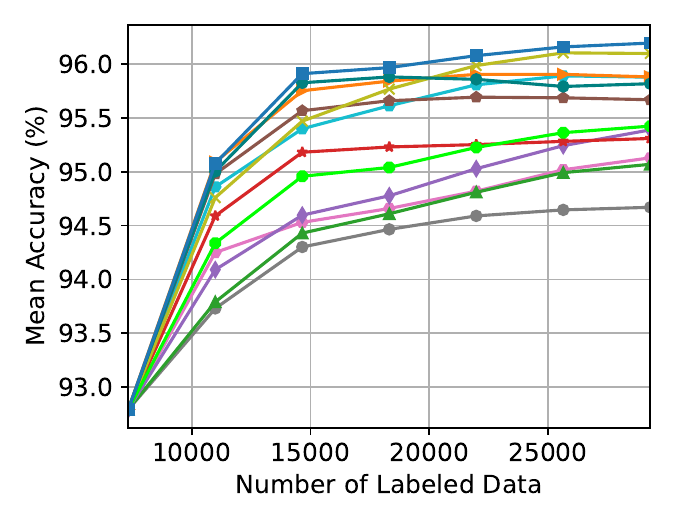}

\includegraphics[width=0.6\linewidth]{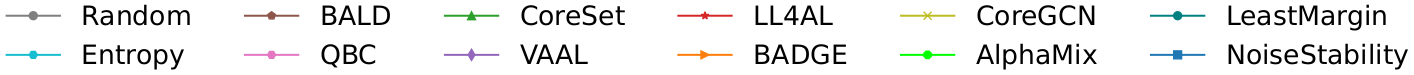}

    \caption{Classification performance due to the different AL methods on MNIST (left), Cifar10 (middle) and SVHN  (right). AUBC with mean and variance is presented in Appendix A.4.3. 
    }
    \label{fig:classification}
\end{figure*}

\section{Experimental Results}
\label{experiment}
We compare our method with the state-of-the-art active learning baselines. Since Random selection is simple yet effective, 
we also include it as a baseline. We firstly validate our method on image classification tasks. 
Then  we conduct extensive experiments on multiple tasks in various domains,
including regression, semantic segmentation, as well as natural language processing. 
The model architectures that we use include CNN, MLP and BERT. We perform 7 to 10 active learning cycles, and the annotation budget for each cycle ranges from 20 to more than 3000. All the data selections are carried without replacement, following the common practice in active learning literature.
All the reported results are averaged over 3 runs for a reliable evaluation, \emph{unless specified otherwise}. For clear comparison, we plot the average accuracy of each cycle in figures, and additionally calculate AUBC (area under the budget curve) with the mean and variance. 
We conduct all the experiments using Pytorch \cite{pytorch} and our source code will be publicly available. 


\subsection{Image Classification}
\label{sec:main_classification}

\noindent
\textbf{Setting.} We use three  classification datasets for evaluations, including MNIST~\cite{lecun1998gradient}, Cifar10 \cite{krizhevsky2009learning} and SVHN \cite{svhn}. MNIST consists of 60000 training samples and we only use a small portion since it is an easy task. The experiments on MNIST are repeated 10 times and we report the averaged results. The Cifar10 dataset includes 50000 training and 10000 testing samples, uniformly distributed across 10 classes. SVHN includes 73257 training and 26032 testing samples, and we do not use the additional training data in SVHN, following the common practice. SVHN is a typical imbalanced dataset. 

We use different architectures for these three tasks, including a small CNN for MNIST, VGG19~\cite{simonyan2014very} for Cifar10 and ResNet-18 \cite{he2016deep} for SVHN. The network and training details are described in Appendix A.4.1. Further, in Appendix A.4.2, we present additional results on Cifar10, Cifar100 and Caltech101 with ResNet-18 \cite{he2016deep}.

\noindent 
\textbf{Observations.} We compare our method with the state-of-the-art baselines and illustrate the results in Figure \ref{fig:classification}. The baseline methods include Random selection, Entropy, Least Margin, BALD~\cite{houlsby2011bayesian,gal2017deep}, QBC \cite{kuo2018cost}, CoreSet \cite{sener2018active}, VAAL \cite{sinha2019variational}, LL4AL \cite{yoo2019learning}, BADGE \cite{ash2019deep}, CoreGCN
 ~\cite{Caramalau_2021_CVPR} and ALFA-Mix \cite{parvaneh2022active}.  For our method, we use the following hyper-parameters in all the experiments: $K=30$ and $\zeta = 10^{-3}$, to ensure the constraint on 
noise magnitude. 
The number of Monte-Carlo sampling for BALD is 50. Other  hyper-parameters for baseline methods are set as suggested in the original papers.

As shown in Figure~\ref{fig:classification}, our method (\textit{NoiseStability}) outperforms or is comparable to the best baselines at all active learning cycles. In particular, our method usually yields a more rapid and stable increase of accuracy at the first several cycles, implying that noise stability is not sensitive to the predictive accuracy and calibration. 
In addition, the superior results in the SVHN (and Caltech101 in Appendix) experiments demonstrate that our method is robust to  imbalanced datasets. We conjecture this is due to the simplicity of our method. In constrast, training auxiliary models \cite{kuo2018cost,sinha2019variational} with imbalanced data could be adverse to the auxiliary models themselves, thus leading to an inferior performance of the task model.

\begin{figure*}[t]

    \centering
    \includegraphics[width=0.33\linewidth]{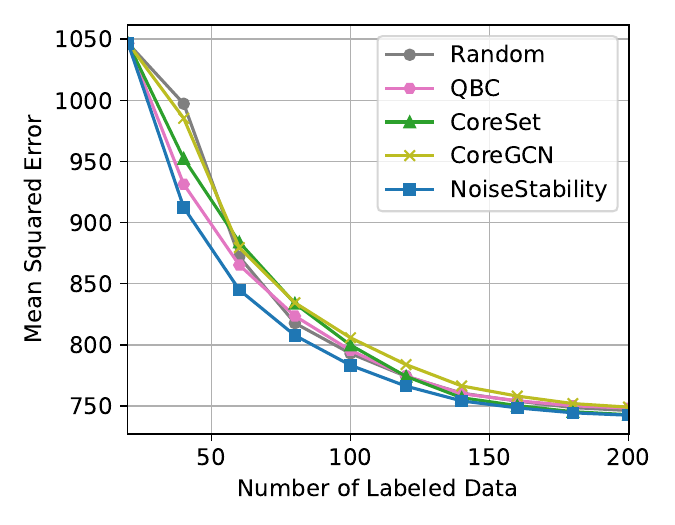}
    \includegraphics[width=0.33\linewidth]{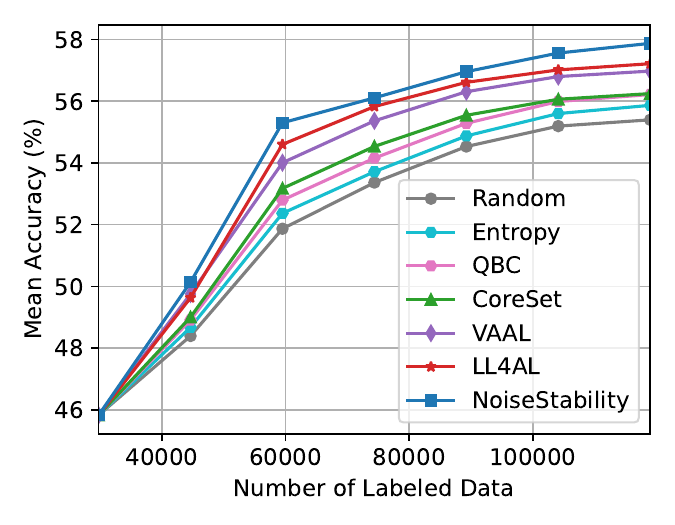}
    \includegraphics[width=0.33\linewidth]{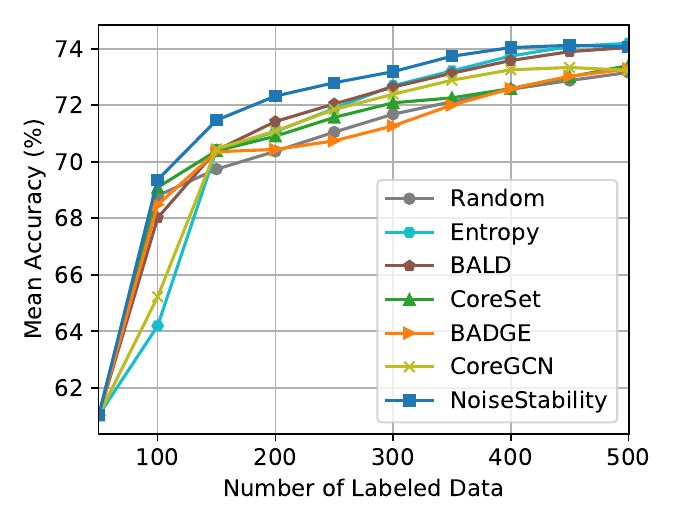}
    \caption{Active learning performance due to the different methods on the regression task: Ames (left), semantic segmentation task: Cityscapes (middle), and natural language processing task: MRPC (right). AUBC results are presented in Appendix A.5.2. }
    \label{fig:more_exp}

\end{figure*}

\subsection{Other Tasks and Data Types}
\label{sec:main_various_tasks}
In this section, we validate our method with more diverse tasks, including regression, semantic segmentation and paraphrase recognition. 
For the regression task, we use data that includes both continuous and discrete variables, and we compare our method with the general algorithms that do not rely on probabilistic predictions. For the semantic segmentation task, we focus on the baselines that have been shown very effective for image data, e.g. LL4AL and VAAL. For the paraphrase recognition task, we use natural language data and make comparisons with the competitive baseline methods designed for classification problems, such as BALD and BADGE. General baselines Random and CoreSet are evaluated on all the three benchmarks. We present a brief result below, and refer readers to Appendix A.5.1 for detailed descriptions of the datasets and experimental configurations.

\subsubsection{Regression Problem}
\label{sec:main_regression}
We use the Ames Housing dataset~\cite{de2011ames} for this task. An MLP model takes a 215-dimensional feature as input and is trained to predict a real value as the house price. Compared with classification, the predictive output of regression does not explicitly contain information about uncertainty, which makes a part of active learning algorithms unavailable. The reported results are averaged over 50 runs for a reliable evaluation. As illustrated in Figure~\ref{fig:more_exp} (left), our method always yields the lowest mean absolute error before the performance saturation. We also observe that, the naive random selection becomes very competitive after the first two cycles, indicating that it is challenging for active learning to outperform passive learning in regression problems.  

\subsubsection{Semantic Segmentation}
Semantic segmentation is more complex than image classification due to the requirement of high classification accuracy at pixel level. As a consequence, in several baseline methods, such as VAAL  \cite{sinha2019variational}, 
the auxiliary models need a careful design in order to achieve an satisfactory performance. This increases the burden of deployment in real scenarios. In contrast, our method is free of extra design and can be easily adopted for this application, demonstrating its task-agnostic advantage. 
We illustrate the results in Figure \ref{fig:more_exp} (middle), and observe a remarkable improvement of our method. Given the Cityscapes dataset is highly imbalanced, this experiment once again demonstrates the superiority of our method on imbalanced datasets. 

\subsubsection{Natural Language Processing}
To further demonstrate the domain-agnostic nature of our method, we conduct an experiment on the Microsoft Research Paraphrase Corpus (MRPC) dataset, a classical NLP task aiming to identify whether two sentences are paraphrases of each other, by fine-tuning a pre-trained BERT model. 
As shown in Figure \ref{fig:more_exp} (right), our method exhibits significant superiority, demonstrating its wide applicability, whereas other state-of-the-art baselines perform less competitive in this task. 



%% file: sections/4_discussion.tex
\section{Discussion}
In this section, we present further discussions on the efficacy of the proposed method, including the choice of hyperparameters, ablation study and computational efficiency. Here we present brief results and put details in Appendix A.6-A.8. 

\subsection{Sensitivity to Hyper-parameters (Relevant to Noise)}
\label{sec:hyperparameter}
We firstly investigate how our algorithm is sensitive to the hyper-parameters (i.e. noise magnitude and sampling times). 
We conduct this set of experiments on MNIST. 

\noindent
\textbf{Noise Magnitude.} An appropriate setting of the noise magnitude will benefit our method. 
Specifically, we observe that noise with a too large magnitude tends to destroy the learned knowledge in DNNs. We evaluate the choices of $\{10^{-6},10^{-4}, 10^{-2}, 1, 10\}$ respectively for the $\zeta$. The results in Appendix A.6 (left) show that a $\zeta$ of between $10^{-4}$ and $10^{-2}$ is likely to yield good performance. A larger $\zeta$ leads to dramatic performance decrease at early cycles. Interestingly, we observe that a too small $\zeta$ also tends to be sub-optimal. We conjecture the reason may lie in that, if the noise is too trivial, only those activated neurons have output perturbations, making the final output deviation less informative.

\noindent
\textbf{Noise Sampling Times.} Accurate approximations such as that for Eq~(\ref{eq:jacobian}) usually require considerable sampling times. However, we surprisingly find that our method 
is not that sensitive to the sampling times $K$. We study the choices of $\{1,3,10,30,50\}$ respectively for $K$. As shown in Appendix A.6 (middle), using $K=10$ can deliver a very competitive performance. 
We conjecture this is because a random noise is sufficient to produce rich pattern perturbations 
for the parameters in DNNs, making the output deviation (due to only a few noise samplings) a reasonable criterion for unlabeled data selection (considering both magnitude and direction).


\noindent
\textbf{Subset Selection.} The topic of subset selection has been well studied in the machine learning community. Other classical algorithms can be easily integrated into our framework. We validate this by replacing our greedy k-center algorithm with kmeans++~\cite{arthur2006k} (suggested by BADGE) and k-DPP~\cite{kulesza2011k}. Results in Appendix A.6 (right) show that the three algorithms yield quite similar results in our framework.


\subsection{Ablation Study}
Here we demonstrate that both the gradient magnitude and direction are informative in subset selection. Specifically, we compare our method with its two degenerated versions including (1) \emph{NoiseStability-M} which selects samples with maximum magnitude $\|\Delta z(\bm{x})\|_2$; and (2) \emph{NoiseStability-D} which normalizes each $\Delta z(\bm{x})$ to unit magnitude before performing the greedy k-center selection. We observe in Appendix A.7 (left) that, neither of the two sub-optimal implementations can compete with the full version of our method. 

We then investigate the effectiveness of the magnitude $\|\Delta z(\bm{x})\|_2$ on uncertainty estimation by evaluating single sample selection. We compare our method with the popular deep Bayesian active learning baselines on the MNIST dataset. For fair comparison, we perform 50 samplings for all the methods. As shown in Appendix A.7 (right), ours is superior to the baselines, demonstrating that noise stability is a reliable indicator of example uncertainty. 

\subsection{Time Efficiency}
The time efficiency of our method depends on the cost of noise sampling and diverse subset selection. With the same hyperparameters as used in experiments, our method exhibits not only comparable computational efficiency among state-of-the-art algorithms, but also good scalability in terms of time complexity, as the annotation budge or category size increases (Appendix A.8).

%% file: sections/6_conclusion.tex
\section{Conclusion}
We propose a simple yet robust method that leverages noise stability for uncertainty estimation in deep active learning. For an input sample, 
noise stability measures the stability of the output 
when a random perturbation is imposed on the model parameters. The direction of the feature deviation is also utilized to ensure the diversity of selected examples. Our theoretical analyses prove that, the resulting deviation under a small parameter noise acts as a reasonable proxy of the gradient, in terms of both magnitude and direction. The extensive evaluations on multiple tasks
demonstrate the superiority of the proposed method.

%% file: sections/9_appendix.tex
\section{Appendix}

\subsection{A.1 Proof and Concentration of Eq (\ref{eq:jacobian})}
\label{sec:proof}

\subsubsection{Proof of Eq (\ref{eq:jacobian})}
\label{sec:proof6}
For brevity, we assume $f(.;\bm{\theta})$ is a real-valued function parameterized by $\bm{\theta} \in \mathbb{R}^n$, and the results can be naturally extended to a vector-valued function by aggregating the result w.r.t. each coordinate of the output vector. 

Specifically, for each $k$, we firstly sample $\bm{t}^{(k)} \in \mathbb{R}^n$ from $\mathcal{N}(\bm{0}, \mathbf{I})$, and then get its direction $\bm{u}^{(k)}$ by self-normalization as $\bm{u}^{(k)} =\frac{ \bm{t}^{(k)}}{\|\bm{t}^{(k)}\|_2}$. Since $\bm{u}^{(k)}$ is uniformly distributed over the unit sphere, the following equation is true:
\begin{equation}
    \mathbb{E} [ \bm{u}^{(k)} {\bm{u}^{(k)}}^T ] = \frac{1}{n} \bm{\mathrm{I}},
\label{eq:uncorrelated}
\end{equation}
To avoid ambiguity, we use $\nabla_{\bm{\theta}} f(\bm{x})$ to denote the gradient in all the following proofs, and omit the notation of the  distribution under the expectation. 

\begin{proof}
By applying Eq (\ref{eq:uncorrelated}), we can expand the inner product between $\nabla_{\bm{\theta}} f(\bm{x})$ and $\bm{u}^{(k)}$ in Eq (\ref{eq:taylor2}) as
\begin{equation}
\begin{aligned}
    \mathbb{E} \|\widehat{\Delta z}(\bm{x})\|_2^2 &= \mathbb{E} \{  \frac{n}{K}  \sum_{k=1}^{K} \|\nabla_{\bm{\theta}} f(\bm{x})^T  \bm{u}^{(k)}\|_2^2 \} \\
    &= \frac{n}{K} \mathbb{E} \{\sum_{k=1}^{K} \nabla_{\bm{\theta}} f(\bm{x})^T \bm{u}^{(k)} {\bm{u}^{(k)}}^T f(\bm{x}) \} \\
    &= \frac{1}{K} \mathbb{E} \{\sum_{k=1}^{K} \nabla_{\bm{\theta}} f(\bm{x})^T f(\bm{x}) \} \\
    &= \|\nabla_{\bm{\theta}} f(\bm{x}) \|_2^2.
\label{eq:j_proof}
\end{aligned}
\end{equation}
When $f(.;\bm{\theta})$ is vector-valued, the above proof can be extended by applying Eq (\ref{eq:j_proof}) to each dimension and then aggregate the result of each dimension by adding them together. We can do this for both sides of Eq (\ref{eq:j_proof}), and the equation still holds.  

\end{proof} 

\subsubsection{Concentration Inequality}
We further present the concentration analysis of the empirical noise stability $\|\widehat{\Delta z}(\bm{x})\|_2^2$. The main result is:

\textbf{Concentration of the gradient magnitude.}
 \textit{Let $\{\bm{u}^{(k)} \in \mathbb{R}^n |k=1,2,...,K\}$ be independent random directions, $\{\bm{x} | \bm{x} \in X_U\}$ be $N$ unlabeled candidates, $\widehat{\Delta z}(\bm{x})$ be obtained by Eq (\ref{eq:main}) and $\nabla_{\bm{\theta}} f(\bm{x}) \in \mathbb{R}^n$ be the gradient for a real-valued function $f$. We assume that $\epsilon > 0$, $c$ and $C$ are constants and $K \geq (C/\epsilon^2) \log N$. Then, with probability at least $1-2 \exp{\{-c \epsilon^2 K\}}$, we have}
 \begin{equation}
 \label{eq:concentration1}
|\|\widehat{\Delta z}(\bm{x})\|_2- \|\nabla_{\bm{\theta}} f(\bm{x})\|_2| \leq \epsilon \|\nabla_{\bm{\theta}} f(\bm{x})\|_2, \forall \bm{x} \in X_U.
 \end{equation}

\begin{proof}
To prove this, we construct a matrix $\bm{Q} \in \mathbb{R}^{K \times n}$ of which $k$-th row $\bm{Q}_k$ is the transpose of $\sqrt{\frac{n}{K}} \bm{u}^{(k)}$.  Then, by substituting the first order Taylor expansion, we have
\begin{equation}
    \widehat{\Delta z}(\bm{x}) = \bm{Q} \nabla_{\bm{\theta}} f(\bm{x}),
\end{equation}
where  $\bm{Q}$ can be regarded as a scale-preserved random projection (in  $\mathbb{R}^n$) onto a low-dimensional space $\mathbb{R}^K$. Then, the above concentration result can be proved by directly applying the Johnson-Lindenstrauss lemma (Theorem 5.3.1 in ~\cite{vershynin2018high}).
\end{proof}


\subsection{A.2 Proof and Concentration for Eq (\ref{eq:property1}-\ref{eq:property2})}
\label{sec:proof2}
\subsubsection{Proof of Eq (\ref{eq:property1}) and Eq (\ref{eq:property2})}
The proof of Eq (\ref{eq:property1}) follows the same key idea of utilizing the property that the elements of the vector $\bm{u}$ are uncorrelated, as shown in Eq (\ref{eq:uncorrelated}). Using a proper scaling, $\bm{Q}$ can be a scale-preserved random projection based on $\{\bm{u}^{(k)}\}_{k=1}^{K}$. Then, Eq (\ref{eq:property1}) can be proved by
\begin{equation}
\begin{aligned}
   \mathbb{E} \|
\widehat{\Delta z}(\bm{x}_i) - \widehat{\Delta z}(\bm{x}_j)\|_2^2 &= \mathbb{E}  \|\bm{Q}[\nabla_{\bm{\theta}}f(\bm{x}_i)-\nabla_{\bm{\theta}}f(\bm{x}_j)]
\|_2^2 \\
&= \|\nabla_{\bm{\theta}}f(\bm{x}_i)-\nabla_{\bm{\theta}}f(\bm{x}_j) \|_2^2.
\end{aligned}
\end{equation}

For proving Eq (\ref{eq:property2}), we expand the quadratic terms on both sides of Eq (\ref{eq:property1}) as
\begin{equation}
\begin{aligned}
\mathbb{E} \|
\widehat{\Delta z}(\bm{x}_i) - \widehat{\Delta z}(\bm{x}_j)\|_2^2 = & \mathbb{E} \|
\widehat{\Delta z}(\bm{x}_i) \|^2_2 + \mathbb{E} \|
\widehat{\Delta z}(\bm{x}_j) \|^2_2 \\ & - 2 \mathbb{E} \langle \widehat{\Delta z}(\bm{x}_i), \widehat{\Delta z}(\bm{x}_j) \rangle, \\
\|\nabla_{\bm{\theta}}f(\bm{x}_i)-\nabla_{\bm{\theta}}f(\bm{x}_j) \|_2^2 =  & \|\nabla_{\bm{\theta}}f(\bm{x}_i)\|^2_2 +  \|\nabla_{\bm{\theta}}f(\bm{x}_j)\|^2_2 \\ & - 2 \langle \nabla_{\bm{\theta}}f(\bm{x}_i), \nabla_{\bm{\theta}}f(\bm{x}_j) \rangle.
\end{aligned}
\end{equation}
Then substituting 
$\mathbb{E} \|\widehat{\Delta z}(\bm{x})\|_2^2$ with  $\|\nabla_{\bm{\theta}} f(\bm{x}) \|_2^2$ (the known fact in Eq (\ref{eq:jacobian})) proves Eq (\ref{eq:property2}).

\subsubsection{Concentration Inequality}

\textbf{Concentration of the Euclidean distance.}
 \textit{Let $\{\bm{u}^{(k)} \in \mathbb{R}^n |k=1,2,...,K\}$ be independent random directions, $\{\bm{x} | \bm{x} \in X_U\}$ be $N$ unlabeled candidates, $\widehat{\Delta z}(\bm{x})$ be obtained by Eq (\ref{eq:main}) and $\nabla_{\bm{\theta}} f(\bm{x}) \in \mathbb{R}^n$ be the gradient for a real-valued function $f$. We assume that $\epsilon > 0$, $c$ and $C$ are constants and $K \geq (C/\epsilon^2) \log N$. Then, with probability at least $1-2 \exp{\{-c \epsilon^2 K\}}$, we have}
 \begin{equation}
 \begin{aligned}
|\|\widehat{\Delta z}(\bm{x}_i)-\widehat{\Delta z}(\bm{x}_j)\|_2- \|\nabla_{\bm{\theta}} f(\bm{x}_i) - \nabla_{\bm{\theta}} f(\bm{x}_j)\|_2| \\ \leq 
      \epsilon \|\nabla_{\bm{\theta}} f(\bm{x}_i) - \nabla_{\bm{\theta}} f(\bm{x}_j)\|_2, \forall \bm{x}_i,\bm{x}_j \in X_U.
\end{aligned}
 \end{equation}
 Note that both the formula and proof are the same as the concentration of the magnitude in Eq (\ref{eq:concentration1}), as long as regarding $\nabla_{\bm{\theta}} f(\bm{x}_i) - \nabla_{\bm{\theta}} f(\bm{x}_j)$ as the original vector and $\bm{Q}$ as the random projection. 

\textbf{Concentration of the inner product.} When considering the inner product, we use  normalized vectors to discuss the concentration property. This is because for pairwise similarity comparison, the angle between two vectors is more important to our analysis than the original inner product.
Denote the normalized inner product of the output deviation by $\hat{S}_{ij}$, as
 \begin{equation}
        \hat{S}_{ij} = 
       \langle \frac{\widehat{\Delta z}(\bm{x}_i)}{\|\widehat{\Delta z}(\bm{x}_i)\|_2}, \frac{\widehat{\Delta z}(\bm{x}_j)}{\|\widehat{\Delta z}(\bm{x}_j)\|_2} \rangle.
 \end{equation}
Our main result is claimed as follows.

 \textit{Let $\{\bm{u}^{(k)} \in \mathbb{R}^n |k=1,2,...,K\}$ be independent random directions, $\{\bm{x} | \bm{x} \in X_U\}$ be $N$ unlabeled candidates, $\widehat{\Delta z}(\bm{x})$ be obtained by Eq (\ref{eq:main}), $\nabla_{\bm{\theta}} f(\bm{x}) \in \mathbb{R}^n$ be the gradient for a real-valued function $f$ and $g(\bm{x}) = \nabla_{\bm{\theta}} f(\bm{x}) / \|\nabla_{\bm{\theta}} f(\bm{x})\|_2$. We assume that $\epsilon > 0$, $c$ and $C$ are constants and $K \geq (C/\epsilon^2) \log N$. Then, with probability at least $1-2 \exp{\{-c \epsilon^2 K\}}$, we have $\forall \bm{x}_i,\bm{x}_j \in X_U$ }
 \begin{equation}
        \frac{\langle g(\bm{x}_i),g(\bm{x}_j) \rangle - \epsilon}{(1+\epsilon)^2}
        \leq
       \langle \hat{S}_{ij} \rangle 
       \leq 
       \frac{\langle g(\bm{x}_i),g(\bm{x}_j) \rangle + \epsilon}{(1-\epsilon)^2}.
\label{eq:concentration3}
 \end{equation}
 \begin{proof}
We further simplify $g(\bm{x}_i), g(\bm{x}_j)$ as $\bm{g}_i, \bm{g}_j$, respectively. Then by applying Eq (\ref{eq:concentration1}) to $\bm{g}_i+\bm{g}_j$ and $\bm{g}_i-\bm{g}_j$, we have that with  probability at least $1-2 \exp{\{-c \epsilon^2 K\}}$, the following two statements are true $\forall \bm{x}_i,\bm{x}_j \in X_U$: 
\begin{equation}
\small
\begin{aligned}
     & (1-\epsilon) \|(\bm{g}_i+\bm{g}_j)\|^2_2 \leq \|\bm{Q}(\bm{g}_i+\bm{g}_j) \|^2_2 \leq (1+\epsilon) \|(\bm{g}_i+\bm{g}_j)\|^2_2, \\
     & (1-\epsilon) \|(\bm{g}_i-\bm{g}_j)\|^2_2 \leq \|\bm{Q}(\bm{g}_i-\bm{g}_j) \|^2_2 \leq (1+\epsilon) \|(\bm{g}_i-\bm{g}_j)\|^2_2. 
\end{aligned}
\end{equation}
Then, we have that $\forall \bm{x}_i,\bm{x}_j \in X_U$, the following derivation is true with the same probability:
\begin{equation}
\begin{aligned}
    & \langle \frac{\widehat{\Delta z}(\bm{x}_i)}{\|\widehat{\Delta z}(\bm{x}_i)\|_2}, \frac{\widehat{\Delta z}(\bm{x}_j)}{\|\widehat{\Delta z}(\bm{x}_j)\|_2} \rangle \\
    = & \langle\ \frac{\bm{Q} \nabla_{\bm{\theta}} f(\bm{x}_i)}{\|\bm{Q} \nabla_{\bm{\theta}} f(\bm{x}_i)\|_2}, \frac{\bm{Q} \nabla_{\bm{\theta}} f(\bm{x}_j)}{\|\bm{Q} \nabla_{\bm{\theta}} f(\bm{x}_j)\|_2} \rangle \\
    = & \frac{ \bm{Q} \langle\ \nabla_{\bm{\theta}} f(\bm{x}_i), \nabla_{\bm{\theta}} f(\bm{x}_j) \rangle}{\|\bm{Q} \nabla_{\bm{\theta}} f(\bm{x}_i)\|_2 \|\bm{Q} \nabla_{\bm{\theta}} f(\bm{x}_j)\|_2} \\
    \geq & \frac{ \bm{Q} \langle\ \nabla_{\bm{\theta}} f(\bm{x}_i), \nabla_{\bm{\theta}} f(\bm{x}_j) \rangle}{(1+\epsilon)^2 \|\nabla_{\bm{\theta}} f(\bm{x}_i)\|_2 \| \nabla_{\bm{\theta}} f(\bm{x}_j)\|_2} \\
    = & \frac{ \bm{Q} \langle\ \bm{g}_i, \bm{g}_j \rangle}{(1+\epsilon)^2} \\
    = & \frac{1}{4(1+\epsilon)^2}(\|\bm{Q}(\bm{g}_i+\bm{g}_j) \|^2_2 - \|\bm{Q}(\bm{g}_i-\bm{g}_j) \|^2_2) \\
     \geq & \frac{1}{4(1+\epsilon)^2}((1-\epsilon)\|\bm{g}_i+\bm{g}_j\|^2_2 - (1+\epsilon)\|\bm{g}_i-\bm{g}_j \|^2_2)\\
    = & \frac{1}{4(1+\epsilon)^2}(4\langle \bm{g}_i,\bm{g}_j \rangle - 2 \epsilon (\|\bm{g}_i\|_2^2 + \|\bm{g}_j \|^2_2))\\
    = & \frac{\langle \bm{g}_i,\bm{g}_j \rangle - \epsilon}{(1+\epsilon)^2}.
\end{aligned}
\end{equation}
The other direction of Eq (\ref{eq:concentration3}) can be similarly derived. 
Combining both directions completes the proof.
 \end{proof}
 
\noindent \textbf{Remark.} In all the three concentration guarantees, we involve the requirement of the noise sampling times $K$ as $K \geq (C/\epsilon^2) \log N$, where $N$ is the number of unlabeled candidates. The 
increment of $K$ is only logarithmically proportional to the increment of $N$,
indicating that noise stability is efficient in sampling. This partly explains why our algorithm is capable of achieving a decent performance with a relatively small $K$ (e.g. $K=10$). 
 

\subsection{A.3 Derivation of Eq (\ref{eq:var_change})}
Given $\mathbf{A}_{S^2}$ defined in the main text, the output variance at the input point $\bm{x}$ can be represented~\cite{alma991005548549706306} by
\begin{equation}
    \sigma^2_{\hat{y}}(\bm{x}) \approx \mathrm{J}_{\bm{\theta}}(\bm{x};\bm{\theta})^T \mathbf{A}_{S^2}^{-1}  \mathrm{J}_{\bm{\theta}}(\bm{x};\bm{\theta}).
\end{equation}
With the normality and local linearity assumptions, the authors in~\citep{MacKayDavidJ.C1992IOFf,CohnDavidA1996NNEU} further quantify the change of output variance. Specifically, when a new sample $\tilde{\bm{x}}$ is labeled and added to the training set, the expected new output variance at $\bm{x}$ is
\begin{equation}
    \tilde{\sigma}^2_{\hat{y}}(\bm{x}) \approx \sigma^2_{\hat{y}}(\bm{x}) - \frac{\sigma^2_{\hat{y}}(\bm{x}, \tilde{\bm{x}})}{S^2 + \sigma^2_{\hat{y}}(\tilde{\bm{x}})}, 
\end{equation}
where $\sigma^2_{\hat{y}}(\bm{x}, \tilde{\bm{x}})$ is defined as
\begin{equation}
    \sigma_{\hat{y}}(\bm{x}, \tilde{\bm{x}}) \equiv  \mathrm{J}_{\bm{\theta}}(\bm{x};\bm{\theta})^T \mathbf{A}_{S^2}^{-1}  \mathrm{J}_{\bm{\theta}}(\tilde{\bm{x}};\bm{\theta}).
\end{equation}
Therefore, by adding a new sample $\tilde{\bm{x}}$, the reduction of the output variance $\sigma^2_{\hat{y}}(\bm{x})$ is:
\begin{equation}
\begin{aligned}
    \underset{\tilde{\bm{x}}}{\Delta} \sigma^2_{\hat{y}}(\bm{x}) &= \frac{\sigma^2_{\hat{y}}(\bm{x}, \tilde{\bm{x}})}{S^2 + \sigma^2_{\hat{y}}(\tilde{\bm{x}})} \\
    &= \frac{\|\mathrm{J}_{\bm{\theta}}(\bm{x};\bm{\theta})^T \mathbf{A}_{S^2}^{-1}  \mathrm{J}_{\bm{\theta}}(\tilde{\bm{x}};\bm{\theta})\|^2}{S^2+\mathrm{J}_{\bm{\theta}}(\tilde{\bm{x}};\bm{\theta})^T \mathbf{A}_{S^2}^{-1}  \mathrm{J}_{\bm{\theta}}(\tilde{\bm{x}};\bm{\theta})} \\
    &= \frac{\|\mathrm{J}_{\bm{\theta}}(\bm{x};\bm{\theta})^T \mathbf{A}_{S^2}^{-1}  \bm{u}(\tilde{\bm{x}})\|^2 \cdot \|\mathrm{J}_{\bm{\theta}}(\tilde{\bm{x}};\bm{\theta})\|^2}{S^2+\| \bm{u}(\tilde{\bm{x}})^T \mathbf{A}_{S^2}^{-1}  \bm{u}(\tilde{\bm{x}}) \| \cdot \|\mathrm{J}_{\bm{\theta}}(\tilde{\bm{x}};\bm{\theta})\|^2},
\label{eq:var_change}
\end{aligned}
\end{equation}

where we use $\bm{u}(\tilde{\bm{x}})$ of the unit length to denote the direction of $\mathbf{J}_{\hat{y}}(\tilde{\bm{x}})$ as  $\bm{u}(\tilde{\bm{x}}) = \frac{\mathrm{J}_{\bm{\theta}}(\tilde{\bm{x}};\bm{\theta})}{\|\mathrm{J}_{\bm{\theta}}(\tilde{\bm{x}};\bm{\theta})\|^2}$.

\subsection{A.4 Appendix for Image Classification}
\label{sec:exp_img_setting}
\subsubsection{A.4.1 Experiment Details.}

We use three different models for the selected benchmark datasets. The architectures and training details are described as follows. 

\noindent \textbf{MNIST.} We use a small CNN with the architecture specified in Table~\ref{table:cnn_arch}. To train this task model, we use the Adam~\cite{kingma2014adam} optimizer with a learning rate of $1\times10^{-3}$ and a  batch size of 96. At each cycle, we train the model for 50 epochs with the labeled set.

\begin{table}[h]
\begin{center}
\begin{tabular}{ccc}
\toprule
Layer & \# of Channels & Spatial Size
\\
\midrule
Input & 1 & $28 \times 28$ \\
Conv2d (1, 32, 5) & 32 & $24 \times 24$  \\
MaxPool2d & 32& $12 \times 12$  \\
ReLU &  32& $12 \times 12$  \\
Conv2d (32, 64, 5) & 64 & $8 \times 8$  \\
MaxPool2d & 64 & $4 \times 4$  \\
ReLU & 64 & $4 \times 4$  \\
Linear (1024, 128) & 128 & -  \\
Linear (128, 10) & - & -  \\
\bottomrule
\end{tabular}
\end{center}
\caption{The network architecture for MNIST classification.}
\label{table:cnn_arch}
\end{table}

\begin{figure*}[htb]
    \centering
    \includegraphics[width=0.3\linewidth]{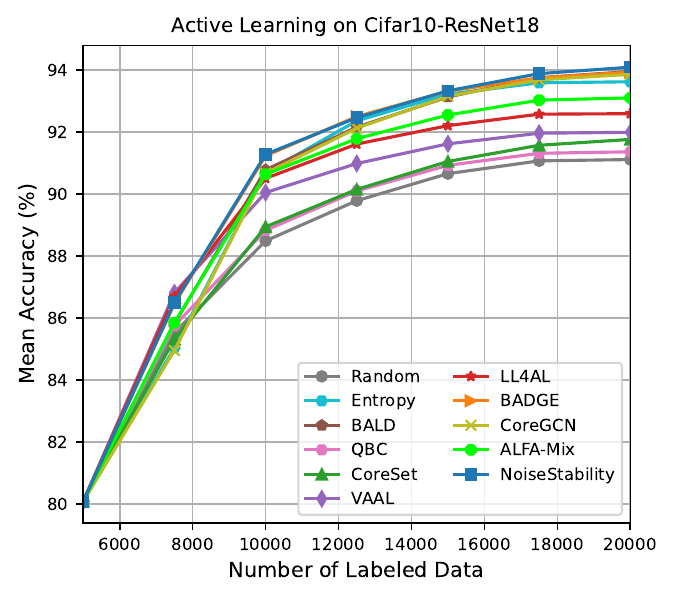}
    \includegraphics[width=0.3\linewidth]{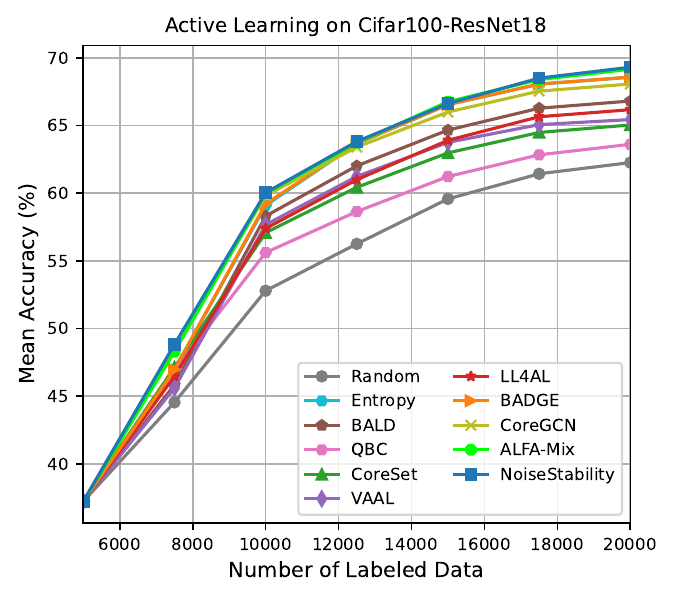}
    \includegraphics[width=0.3\linewidth]{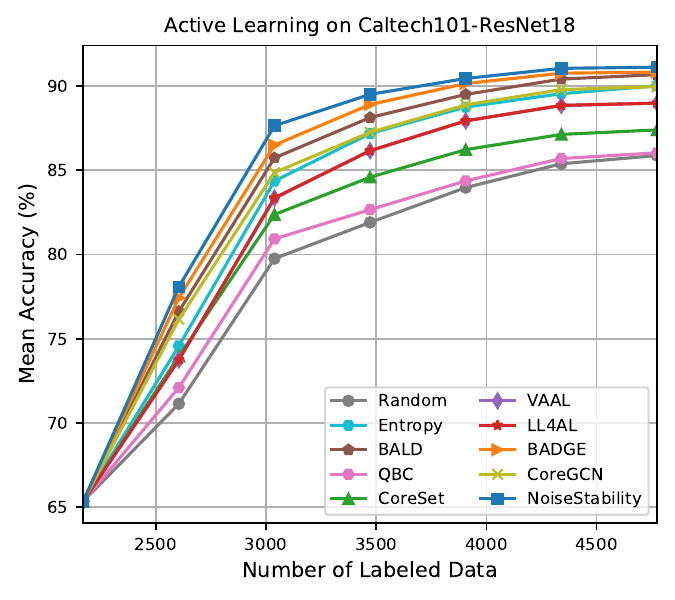}
    \caption{Classification performance due to the different active learning methods with ResNet-18 on Cifar10 (\textbf{left}), Cifar100 (\textbf{middle}) and Caltech101 (\textbf{right}). }
    \label{fig:exp_resnet}
\end{figure*}

\noindent \textbf{Cifar10.} For the Cifar10 task which is more challenging, we use VGG16~\cite{simonyan2014very} as the task model. Specifically, we follow the practice in  \cite{Caramalau_2021_CVPR, zhang2020state} to adopt a special version\footnote{https://github.com/chengyangfu/pytorch-vgg-cifar10} of VGG19 to make the model compatible with the lower image resolution (i.e. $32\times32$) in Cifar10. We use the SGD \citep{bottou2010large} optimizer with a batch size of 128, a momentum of 0.9, a weight decay rate of $5\times10^{-4}$ and an initial learning rate of 0.05. We train the model for 200 epochs and the learning rate is decayed by 0.5 every 30 epochs. 

\noindent \textbf{SVHN.} We use ResNet-18 \cite{he2016deep} as the task model for this experiment. Similar to the Cifar10 experiment, we adopt a special version\footnote{https://github.com/kuangliu/pytorch-cifar} to make the model compatible with the lower image resolution of $32 \times 32$. At each cycle we use a batch size of 128 samples to train the task model with an initial learning rate of 0.1. The training continues for 200 epochs and we decay the learning rate by 0.1
at the $160^{th}$ epoch. We employ the SGD \citep{bottou2010large} optimizer with a momentum of 0.9 and a weight decay rate of $5\times10^{-4}$.

\subsubsection{A.4.2 Additional Experiments.}
\label{sec:exp_img_more}
We conduct extensive experiments using the ResNet-18 architecture, demonstrating the superiority of our algorithm for larger selection sizes (e.g. 2500) and higher image resolutions (e.g. $224 \times 224$). 

\noindent \textbf{Settings.} We use Cifar10~\cite{krizhevsky2009learning}, Cifar100~\cite{krizhevsky2009learning} and Caltech101~\cite{caltech} for this group of experiments. The Cifar100 dataset includes 50000 training and 10000 testing samples, uniformly distributed across 100 classes. The Caltech101 dataset consists of 8677 images of higher resolution (e.g. $300\times200$ pixels), belonging to 101 categories. Caltech101 is a  typical imbalanced dataset, where the number of the samples in a class varies from 40 to 800. For Cifar10 and Cifar100, we use the same settings as that for SVHN. For Caltech101, we use the original ResNet-18 and train the model for 50 epochs at each cycle. The initial learning rate is set to 0.01, decayed by 0.1 at the $40^{th}$ epoch. Due to the higher image resolution, a smaller batch size of 64 is used to avoid GPU memory issues. We use a momentum of 0.9 and a weight decay rate of $5\times10^{-4}$ with the SGD optimizer. For all the three datasets, during training we adopt standard data augmentations including random flip and crop. For Cifar100 and Caltech101, we set the sampling time to $K=10$. 

\noindent \textbf{Observations.} We compare our algorithm with the baselines as mentioned in the main text. Similarly, as shown in Figure~\ref{fig:exp_resnet}, our algorithm exhibits notable superiority. 
Specifically, we observe that on standard balanced datasets Cifar10 and Cifar100, our NoiseStability is slightly leading the most competitive baselines, including Entropy,  BADGE~\cite{ash2019deep} and  CoreGCN~\cite{Caramalau_2021_CVPR}. On Caltech101, our method yields remarkable improvements over the baselines. 

Note that our observations are aligned with the conclusions in the recent studies~\cite{beck2021effective,lang2021best}, which demonstrate that when data augmentation is used, advanced AL algorithms do not significantly outperform simple uncertainty-based methods (e.g. Entropy) on standard benchmarks. Nevertheless, advanced AL algorithms are more competitive on challenging datasets (e.g. imbalanced). 

\subsubsection{A.4.3 Overall Comparison.}
We use the area under the budget curve (AUBC)~\cite{zhan2021comparative} to evaluate the overall performance of each active learning algorithm. Specifically, we report the mean AUBC and its corresponding standard deviation among all trials. The AUBC results for Figure~\ref{fig:classification} are presented in Table~\ref{tab:AUBC_img_cls}. For an easier comparison, we also present the average rankings in Table~\ref{tab:AR_img_cls}.

\begin{table}[tbh]
  \caption{AUBC for image classification tasks.}
  \label{tab:AUBC_img_cls}
  \centering
  \footnotesize
  \begin{tabular}{lccc}
  \toprule
Method & MNIST & Cifar10 & SVHN \\
\midrule
Random & 79.73$\pm$1.00 & 63.46$\pm$1.42 & 94.17$\pm$0.03 \\
Entropy & 79.87$\pm$2.61 & 62.98$\pm$0.81 & 95.18$\pm$0.04 \\ 
Least Margin & 82.50$\pm$1.45 & 61.30$\pm$0.87 & 95.28$\pm$0.03 \\
BALD & 80.49$\pm$2.14 & 58.58$\pm$0.81  & 95.15$\pm$0.01 \\
QBC & 79.89$\pm$2.73 & 61.19$\pm$0.61 & 94.46$\pm$0.13  \\ 
CoreSet & 79.83$\pm$1.82 & 60.66$\pm$0.45 & 94.36$\pm$0.11 \\ 
VAAL & 77.00$\pm$1.41 & 63.66$\pm$1.11  & 94.56$\pm$0.15 \\
LL4AL & 79.33$\pm$1.89 & 63.06$\pm$3.16& 94.81$\pm$0.02 \\
BADGE & 82.22$\pm$1.13 & 63.28$\pm$2.47  & 95.31$\pm$0.01 \\
CoreGCN & 80.24$\pm$1.54 & 62.55$\pm$1.41  & - \\ 
SRAAL & - & - & 95.28$\pm$0.05 \\ 
ALFA-Mix & 83.48$\pm$0.70 & 62.80$\pm$2.92 & 94.73$\pm$0.08 \\
NoiseStability & \textbf{83.00$\pm$0.79} & \textbf{64.46$\pm$0.62} & \textbf{95.46$\pm$0.05} \\
\bottomrule  
  \end{tabular}
\end{table}

\begin{table}[tb]
  \caption{Average rankings for image classification tasks.}
  \label{tab:AR_img_cls}
  \centering
  \footnotesize
  \begin{tabular}{lccc}
  \toprule
Method & MNIST & Cifar10 & SVHN \\
\midrule
Random          & 8.33 & 3.50 & 11.00   \\
Entropy         & 6.78 & 6.67 & 4.33   \\
Least Margin    & 3.22 &     8.33 & 3.16   \\
BALD            & 4.89 & 11.00 & 4.17   \\
QBC             & 6.78 & 9.00 & 8.83   \\
CoreSet         & 7.33 & 9.50 & 10.00   \\
VAAL            & 11.00 & 4.83 & 8.00   \\
LL4AL           & 8.56 & 4.67 & 6.50   \\
BADGE           & 3.22 & 3.50 & 2.33   \\
CoreGCN         & 6.11 & 5.83 & -   \\
SRAAL           & - & - & 3.00   \\
ALFA-Mix        & \textbf{1.11} & 6.33 & 6.67   \\
NoiseStability  & 1.89 & \textbf{1.17} & \textbf{1.17}  \\
\bottomrule  
  \end{tabular}
\end{table}

\subsection{A.5 Appendix For Other Tasks}

\subsubsection{A.5.1 Details of Datasets and Learning Settings.}
\label{sec:exp_other_datasets}
\paragraph{Regression}
\textbf{Dataset.} The Ames Housing dataset~\cite{de2011ames} describes the sales of individual residential properties in Ames, Iowa from 2006 to 2010. It includes 1461 labeled training samples and 60 explanatory attributes involved in assessing home values. This dataset is considered as an extended version of the classical Boston Housing dataset, which only has 506 labeled samples and 14 attributes.

\noindent \textbf{Model and Training.} We use 50\% of the data for training and the remaining for testing. We follow common practices to transform the categorical features into numerical ones with one-hot encoding. A four-layer MLP is employed to extract features of the input vectors, and another fully connected layer is used as the predictor. 

The Adam optimizer is adopted to train the task model. The training will continue for 500 epochs and the learning rate is set to 0.001. For our method, the output deviation w.r.t. a parameter noise is calculated on the final feature rather than the regression output. 

\paragraph{Semantic Segmentation}
\textbf{Dataset.}
We use the Cityscapes \cite{cityscapes}, a large-scale street scene dataset. It includes images taken under various weather conditions in different seasons from 50 cities. 
Following widely used settings, we only use the standard training and validation data and crop the images to a dimension of $688\times688$. There are 19 categories (classes) in total for each pixel, leading to a pixel-level classification problem. It is worth noting that the cropping size also depends on the task model. 

\noindent
\textbf{Model and Training.}
\label{modelselect-2}
Following a common practice \cite{zhang2020state},
we choose a dilated residual network (DRN-D-22) \citep{yu2017dilated} as the task model. It includes 8 layer modules and each module is composed of 1 or 2 \textit{BasicBlock} of layers. One can refer to~\citep{yu2017dilated} for the network details .

Similar to the classification evaluations, we train the model for 7 cycles and 50 epochs for each cycle. Following the suggestion in \citep{yu2017dilated}, we use the Adam optimizer \citep{kingma2014adam} with a learning rate of $5\times10^{-4}$, which remains unchanged during the entire training. 

\paragraph{Natural Language Processing}
\textbf{Dataset.}
The Microsoft Research Paraphrase Corpus (MRPC) dataset ~\cite{dolan2005automatically} is a corpus of sentence pairs for recognizing whether the two sentences in each pair are semantically equivalent. It includes 3669 training samples and two thirds of them are positive. In general, this task can be treated as a binary classification one. 

\noindent \textbf{Model and Training.}
At each cycle, we fine-tune the standard BERT (bert-base-uncased) model for 10 epochs to ensure convergence. We follow the official guide of \emph{Pytorch-Transformers-HuggingFace} to setup the other hyper-parameters. We use 50 initial labeled samples for training. At each cycle, 50 additional samples are selected for annotation.

\subsubsection{A.5.2 Overall Comparision.}
The AUBC results for Figure~\ref{fig:more_exp} are presented in Table~\ref{tab:AUBC_others}. Experiments on Cityscapes are conducted only one time due to the high training cost of semantic segmentation tasks. 

\begin{table}[tbh]
  \caption{AUBC for other tasks. N/A refers to not applicable or requiring carefully additional design (e.g. new architectures) for adaptation.  }
  \label{tab:AUBC_others}
  \centering
  \footnotesize
  \begin{tabular}{lccc}
  \toprule
Method & Housing & Cityscapes & MRPC \\
\midrule
Random & 849.45$\pm$27.56 & 52.08 & 69.96$\pm$3.28\\
Entropy & N/A & 52.42 & 70.26$\pm$4.11 \\ 
QBC & 840.60$\pm$18.82 & 52.74 & - \\
BALD & N/A & N/A & 70.56$\pm$3.68 \\
CoreSet & 847.52$\pm$18.83 & 52.91 & 70.30$\pm$3.32  \\ 
VAAL & N/A & 53.58 & N/A \\
LL4AL & N/A & 53.82 & N/A \\ 
BADGE & N/A & N/A & 70.01$\pm$3.27 \\
CoreGCN & 853.67$\pm$27.93 & - & 70.18$\pm$3.81\\ 
NoiseStability & \textbf{828.78$\pm$16.92} & \textbf{54.25} & \textbf{71.26$\pm$3.74} \\
\bottomrule  
  \end{tabular}
\end{table}

\subsection{A.6 Sensitivity to Hyper-parameters}
\begin{figure*}[t]
    \centering
    \includegraphics[width=0.3\linewidth]{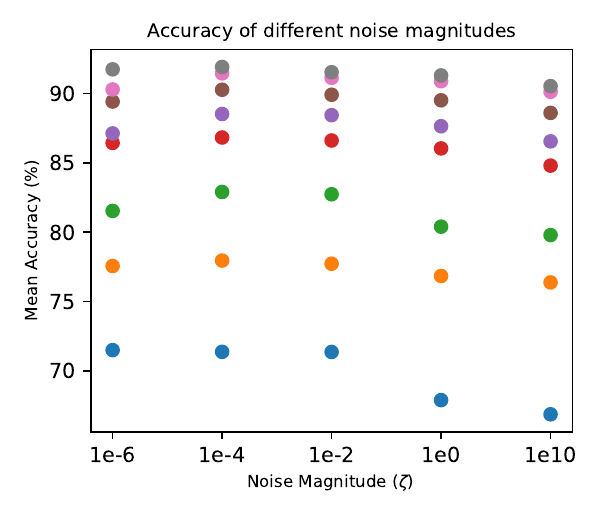}
    \includegraphics[width=0.3\linewidth]{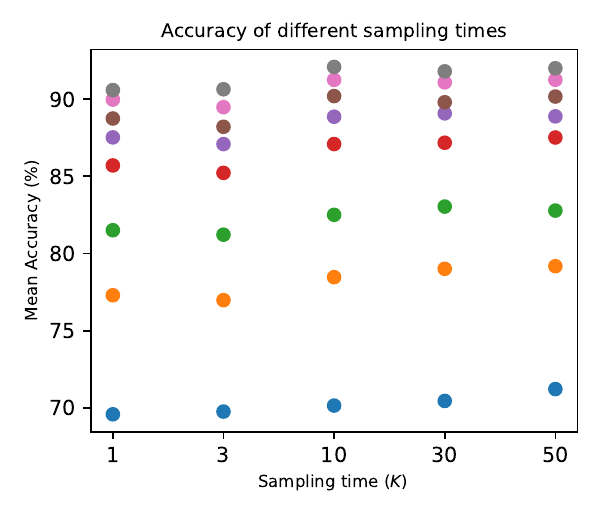}
    \includegraphics[width=0.3\linewidth]{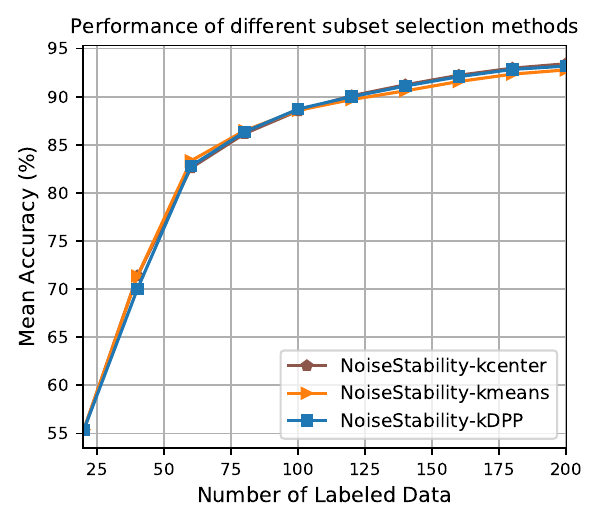}
    \caption{Evaluation of the different hyper-parameters on MNIST: noise magnitude $\lambda$ (\textbf{left}), sampling times $K$ (\textbf{middle}) and subset selection methods (\textbf{right}). In the left and middle plots, we use dots with different colors to denote different cycles, e.g. the blue dots at the bottom refer to cycle 2 and the grey dots at the top refer to cycle 9. }
    \label{fig:discuss_hp}
\end{figure*}

Figure~\ref{fig:discuss_hp} demonstrates details about how the performance of NoiseStability changes when choosing different noise magnitude $\lambda$, sampling times $K$ and subset selection methods.

\subsection{A.7 Ablation Study}
\begin{figure*}[tt]
\centering
    \includegraphics[width=0.4\linewidth]{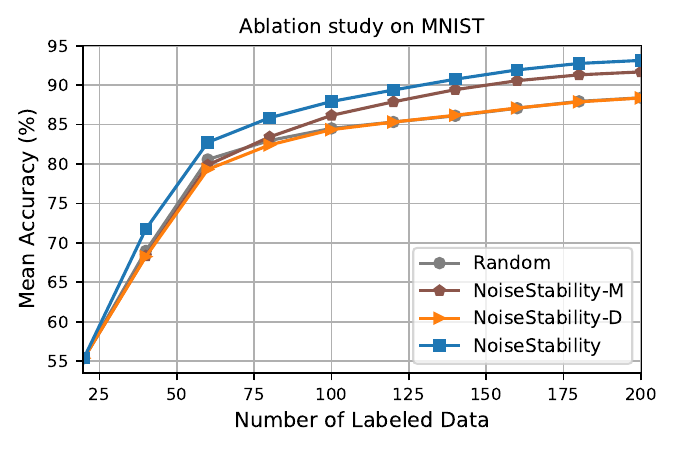}
    \includegraphics[width=0.4\linewidth]{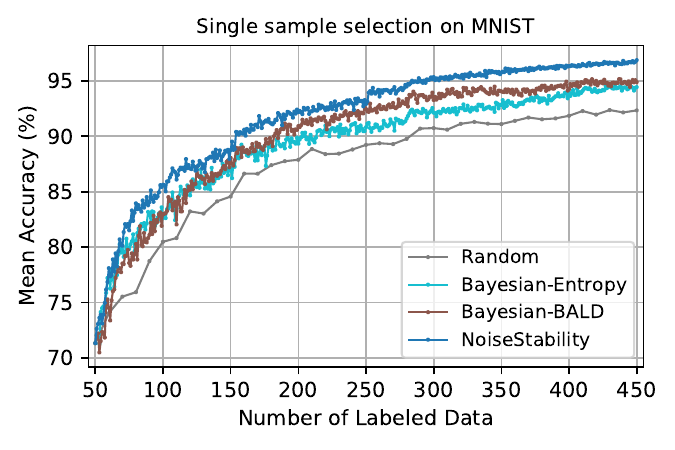}
    \caption{Comparison on MNIST with degenerated implementations (\textbf{left}) and with the Bayesian active learning methods with a selection size of 1 (\textbf{right}). }
    \label{fig:ablation}
\end{figure*}

Figure~\ref{fig:ablation} presents results of ablation study, specifically the individual contribution of the noise magnitude and direction. In addition to arguments in the main paper, we observe from the left figure that, considering only the noise direction without favoring uncertain samples performs almost the same with random selection. However, their joint effect is remarkably more powerful than the sum of their individual effect. 

\subsection{A.8 Time Efficiency}
We conduct two groups of experiments to evaluate time efficiency of active learning methods. In the first group we perform data selection on Cifar10 with ResNet-18, using activ learning settings described in Appendix A4.2. The second group uses Cifar100 with a larger unlabeled pool and acquisition size, as well as the category number. The detailed configurations are listed in Table~\ref{tab:time_config}. Results are presented in Table~\ref{tab:time_result}. 

\begin{table}[h]
  \caption{Experiment settings for evaluating time efficiency.}
  \label{tab:time_config}
  \centering
  \begin{tabular}{lll}
  \toprule
Config & Group1 & Group2 \\
\midrule
dataset & Cifar10 & Cifar100 \\
category number & 10 & 100 \\
unlabeled pool size & 10000 & 25000 \\
acquisition size & 1000 & 5000 \\
\bottomrule  
  \end{tabular}
\end{table}

\begin{table}[h]
  \caption{Running time (minutes) at the middle cycle of different active learning algorithms.}
  \label{tab:time_result}
  \centering
  \begin{tabular}{lll}
  \toprule
Method & Group1 & Group2 \\
\midrule
BALD & 10.8 & 25.1 \\
CoreSet & 4.7 & 13.6 \\
QBC & 34.7 & 35.1 \\
VAAL & 365.1 & 401.4 \\
CoreGCN & 4.6 & 9.9 \\
BADGE & 4.8 & 236.2 \\
NoiseStability & 10.1 & 21.5 \\
\bottomrule  
  \end{tabular}
\end{table}